\documentclass[conference]{IEEEtran}
\usepackage{amsmath,amssymb,amsfonts}
\usepackage{algorithmic}
\usepackage{graphicx}
\usepackage{textcomp}
\usepackage{xcolor}
\usepackage{cite}
\usepackage{hyperref}
\usepackage{booktabs}
\usepackage{amssymb}
\usepackage{multirow}
\usepackage{pifont}
\newcommand{\xmark}{\ding{55}} 
\usepackage{graphicx}

\def\BibTeX{{\rm B\kern-.05em{\sc i\kern-.025em b}\kern-.08em
    T\kern-.1667em\lower.7ex\hbox{E}\kern-.125emX}}
\begin{document}

\title{Demographic Attributes Prediction from Speech Using WavLM Embeddings}

\author{\IEEEauthorblockN{1\textsuperscript{st} Yuchen Yang}
\IEEEauthorblockA{\textit{Dept. of Applied Mathematics and Statistics}\\
\textit{CLSP}\\
\textit{Johns Hopkins University}\\
Baltimore, MD, USA \\
0009-0001-7697-5184}
\and
\IEEEauthorblockN{2\textsuperscript{nd} Thomas Thebaud}
\IEEEauthorblockA{\textit{Dept. of ECE}\\
\textit{CLSP}\\
\textit{Johns Hopkins University}\\
Baltimore, MD, USA \\
0000-0001-8953-7872}
\and
\IEEEauthorblockN{3\textsuperscript{rd} Najim Dehak}
\IEEEauthorblockA{\textit{Dept. of ECE}\\
\textit{CLSP}\\
\textit{Johns Hopkins University}\\
Baltimore, MD, USA \\
0000-0002-4489-5753}
}

\maketitle

\begin{abstract} 
 This paper introduces a general classifier based on WavLM features, to infer demographic characteristics, such as age, gender, native language, education, and country, from speech.
Demographic feature prediction plays a crucial role in applications like language learning, accessibility, and digital forensics, enabling more personalized and inclusive technologies. Leveraging pretrained models for embedding extraction, the proposed framework identifies key acoustic and linguistic features associated with demographic attributes, achieving a Mean Absolute Error (MAE) of 4.94 for age prediction and over 99.81\% accuracy for gender classification across various datasets.
Our system improves upon existing models by up to relative 30\% in MAE and up to relative 10\% in accuracy and F1 scores across tasks, leveraging a diverse range of datasets and large pretrained models to ensure robustness and generalizability. This study offers new insights into speaker diversity and provides a strong foundation for future research in speech-based demographic profiling.
\end{abstract}


\vspace{-5mm}
\section{Introduction}
\label{introduction}
Demographic insights derived from speech data, such as a speaker's age, gender, native language, education, and country, provide valuable knowledge for applications in tailored services, accessibility, and sociolinguistics \cite{ding2020autospeech}. 
By predicting these attributes, technology can adapt to diverse user demographics, enabling more fairness in speech systems without touching the identity of a specific speaker.

Speech contains a wealth of information, not only linguistic content but also cues about a speaker's affective state \cite{thebaud2024multimodal}, health \cite{horiguchi2020endtoend}, and personal characteristics \cite{ding2020autospeech}. Leveraging these non-linguistic features for demographic prediction opens up opportunities for robust, generalized models that capture diverse speaker traits.

This work proposes a novel approach to demographic attribute prediction by:
\begin{enumerate}
    \item Leveraging pretrained foundational models, such as WavLM \cite{baevski2020wav2vec}, for embedding extraction.
    \item Combining multiple datasets with diverse demographic attributes to improve generalization and robustness.
    \item Evaluating cross-dataset performance to demonstrate the adaptability of the approach across varied linguistic and demographic contexts.
\end{enumerate}

The framework uses pretrained embedding extraction models to generate high-dimensional, speaker-specific representations, which are then fed into task-specific classification and regression heads. For age prediction (a regression task), and categorical attributes like gender and native language (classification tasks), we employ Multilayer Perceptron (MLP) \cite{amari1967adaptive}, ResNet32 \cite{szegedy2016inception}, and Long Short-Term Memory (LSTM) \cite{hochreiter1997lstm} architectures. These serve as the classification or regression heads of the overall model, building upon the frozen embeddings provided by the pretrained models. The MLP architecture consists of three layers, while ResNet32 leverages residual connections for deep feature propagation, and LSTM models sequential patterns, making it ideal for tasks involving time-dependent speech features.

By integrating foundational models and diverse datasets, this approach achieves commendable performance on demographic attribute prediction tasks, paving the way for robust cross-dataset generalization.

Section \ref{related_work} reviews previous works on pretrained models for embedding extraction and studies relevant to speaker attribute prediction \cite{wavlm, ding2020autospeech, baevski2020wav2vec}. 
Section \ref{dataset} describes the datasets employed, covering diverse demographic attributes \cite{timit, voxceleb, l2arctic, speech_accent_archive, common_voice}. 
Section \ref{metrics} outlines the evaluation metrics, including Mean Absolute Error (MAE) for age prediction and accuracy and F1 scores for categorical characteristics. 
Section \ref{experiment} explains the experimental workflow, including feature extraction, model architectures, and training and evaluation methods. 
Section \ref{results} presents results, comparing the proposed approach to state-of-the-art methodologies and assessing its effectiveness in representing demographic diversity.

This research introduces a framework for demographic profiling of speech data, leveraging feature extraction methods and WavLM pretrained models \cite{wavlm}. 
The main contributions of this work are as follows:

\begin{itemize}


    \item \textbf{Adaptation of WavLM Embeddings for Demographic Prediction:} While WavLM \cite{wavlm} has been used to improve the state-of-the-art on many speaker-related tasks, such as emotion recognition and speaker recognition, we propose a method to predict speaker characteristics using its features.

    \item \textbf{Cross-Dataset Benchmarking:} We perform a detailed evaluation on five diverse datasets (see Table \ref{tab:data}) to show the model's generalization across demographic attributes \cite{timit, voxceleb, l2arctic, speech_accent_archive, common_voice}.

    \item \textbf{Improvement of the age and gender prediction baseline:} Our study evaluates how well the embeddings capture speaker's characteristics, showing consistent performance improvements for both age and gender.
    
\end{itemize}

\section{Related Works}
\label{related_work}

Recent advancements in deep learning, particularly self-supervised models, have revolutionized speaker attribute prediction by providing high-dimensional embeddings that capture diverse speech characteristics \cite{baevski2020wav2vec}. These embeddings are increasingly employed for all speech tasks, including speaker recognition.

\subsection{Foundational Models for Attribute Prediction}
Foundational models, such as WavLM, leverage self-supervised learning to extract intricate patterns from speech without requiring labeled data. These models produce embeddings that are versatile for various downstream tasks, including demographic attribute prediction.

Based on the Wav2Vec 2.0 framework \cite{baevski2020wav2vec}, WavLM is a self-supervised model trained on diverse datasets to capture acoustic and speaker-specific features. Its embeddings excel in speech emotion recognition~\cite{wu2024emo} and speaker recognition tasks~\cite{ashihara2024investigation}.



\subsection{Speaker Attribute Prediction} 
Speaker attribute prediction has traditionally relied on handcrafted features or statistical models such as i-vectors \cite{dehak2010front}, which represent compact speaker-specific embeddings.
More recently, the statistical approaches were supplanted by the neural approaches such as x-vectors \cite{snyder2018x}.

Recent work has shifted towards deep neural network-based methods. Studies like those by Kwasny et al. \cite{kwasny2020agegender} and Hechmi et al. \cite{hechmi2021voxceleb} demonstrate that embeddings from pretrained models, such as WavLM and Wav2Vec, significantly outperform traditional approaches. These self-supervised embeddings capture general-purpose representations, enabling robust demographic profiling without task-specific feature engineering.

Speaker attribute prediction now primarily uses deep neural networks to predict demographics such as age and gender\cite{almomani2023age}. 
Kwasny et al. \cite{kwasny2020agegender} and Hechmi et al. \cite{hechmi2021voxceleb} illustrate that embeddings from pre-trained speaker verification models can achieve high scores on age and gender classification tasks. 

Recent advances in x-vector-based techniques have given speaker analysis a new layer of analysis for prosody, dialect, and speaker identity. Such embeddings capture high-dimensional data that eliminates the need for tedious feature engineering. For example, traditional features like MFCCs require extensive preprocessing and domain expertise to extract meaningful information, while self-supervised embeddings, like those of Oord et al. \cite{oord2018cpc}, inherently encode rich speaker-specific attributes. Comparative experiments have shown that self-supervised embeddings frequently outperform manual feature sets for demographic profiling purposes \cite{ericsson2021selfsupervised}.


\section{Experimental Pipeline} 
\label{experiment}
In this section, we present the datasets, metrics, pretrained models and architecture choices, as well as the experiments performed.

\subsection{Datasets}
\label{dataset} 

We use five different datasets, each offering a broad range of demographic variables and linguistic contexts. Table \ref{tab:data} shows the distribution of speakers and audio segments within the development and test sets of each dataset.
All datasets are sampled at 16kHz.

\begin{itemize}
\item \textbf{TIMIT} \cite{timit} - The TIMIT dataset is a corpus for phoneme alignment and speech recognition tasks which contains recordings from 630 speakers. The corpus includes 6,300 utterances, where each speaker recites the same set of sentences. These recordings are accompanied by time-aligned word and phoneme transcriptions, enabling precise phonetic analysis. Additionally, the dataset includes an Edu column with 5 categories: 'BS' (Bachelor’s), 'HS' (High School), 'MS' (Master’s), 'PHD' (Doctorate), and 'AS' (Associate).

\item \textbf{VoxCeleb2} \cite{voxceleb} - VoxCeleb2 contains over 1 million utterances from 5,994 speakers, sourced from publicly available YouTube videos. The dataset spans a wide variety of accents and recording conditions, making it suitable for speaker recognition tasks. While VoxCeleb2 does not include explicit demographic labels such as age, demographic data can be inferred or cross-referenced from external sources. To compare with established baselines, we follow the annotations proposed by Hechmi et al. \cite{hechmi2021voxceleb}, using wikipedia-extracted information for the speakers.

\item \textbf{L2Arctic} \cite{l2arctic} - The L2-Arctic dataset consists of speech from 24 non-native L2 English speakers with with 6 native language categories: Arabic, Hindi, Korean, Mandarin, Spanish, and Vietnamese. Each speaker records phonetically balanced English sentences designed to capture a wide range of phoneme combinations. The dataset also includes annotations such as phoneme-level alignment and acoustic features, making it a valuable resource for analyzing non-native pronunciation and accent variability. 

\item \textbf{Speech Accent Archive} \cite{speech_accent_archive} - This dataset consists of recordings from 2140 native and non-native english speakers of 141 countries of origin and 202 native language categories, reading a common English paragraph. This design provides a standardized basis for comparing accent characteristics across origins. The Speech Accent Archive contains metadata about the speakers' native language, age, and gender, which supports studies on linguistic and demographic profiling.

\item \textbf{Common Voice 6.1 (English)} \cite{common_voice} - Common Voice is an open-source collection of speech recordings contributed by volunteers worldwide. The dataset provides metadata fields such as age, gender, and accent; however, this information is optional, and some entries are incomplete or imprecise (e.g., age groups are reported in 10-year ranges). For consistency with other datasets and baseline evaluations, we use only the English subset of version 6.1, including recordings presenting demographic information. This subset includes 3,995 speakers from 17 countries, although the demographic data may not always be fully representative.
\end{itemize}

Table \ref{tab:data} presents an overview of the datasets used for attribute prediction, including the number of speakers, segments, and total recording lengths (in hours) for both development and test sets. The datasets vary significantly in size, from larger corpora like VoxCeleb2 and Common Voice (English), which contain thousands of speakers and hundreds of hours of recordings, to smaller datasets such as L2Arctic, which focuses on a much larger variety of non-native English speakers.

The variation in linguistic and demographic characteristics across these datasets enhances the model’s ability to generalize and capture diverse speaker attributes, supporting demographic profiling across a range of speaker backgrounds.

\begin{table*}[ht]
    \centering
    \caption{Distribution of datasets used.}
    \resizebox{\linewidth}{!}{%
    \begin{tabular}{l c c c c c c c c c c c}
    \toprule
    Dataset & \multicolumn{2}{c}{Speakers} & \multicolumn{2}{c}{Segments} & \multicolumn{2}{c}{Length (hours)} & Age & Gender & Native Lang(\#) & Country(\#) & Edu(\#) \\
    \cmidrule(lr){2-3} \cmidrule(lr){4-5} \cmidrule(lr){6-7} \cmidrule(lr){8-12}
    & Dev & Test & Dev & Test & Dev & Test & & & & & \\
    \midrule
    Speech Accent Archive & 1712 & 428 & 1712 & 428 & 13.24 & 3.21 & \checkmark & \checkmark & \checkmark(202) & \checkmark(141) & \xmark \\
    TIMIT & 461 & 168 & 4610 & 1680 & 3.93 & 1.44 & \checkmark & \checkmark & \xmark & \xmark & \checkmark(5) \\
    VoxCeleb2 & 3680 & 84 & 106922 & 2713 & 233.06 & 5.82 & \checkmark & \checkmark & \xmark & \xmark & \xmark \\
    L2Arctic & 19 & 5 & 21212 & 5655 & 21.17 & 5.90 & \xmark & \checkmark & \checkmark(6) & \xmark & \xmark \\
    Common Voice 6.1 (English) & 3158 & 837 & 127436 & 1368 & 195.38 & 2.15 & \xmark & \checkmark & \xmark & \checkmark(17) & \xmark \\
    \bottomrule
    \end{tabular}%
    }
    \label{tab:data}
\end{table*}

\subsection{Metrics}
\label{metrics} 
The model's performance is assessed using three primary metrics for age prediction and categorical attribute classification.

\begin{itemize} \item \textbf{Mean Absolute Error (MAE):}  In age prediction, a continuous variable, MAE quantifies the mean of the absolute discrepancies between projected and actual age values. It provides a straightforward measure of prediction error without heavily penalizing outliers. MAE is formally defined as: \begin{equation} 
\text{MAE} = \frac{1}{n} \sum_{i=1}^{n} |y_i - \hat{y}_i| 
\end{equation} 

\item \textbf{Accuracy and F1 Score:} Accuracy and F1 Score are utilized for categorical factors such as gender, native language, education, and nationality. Accuracy evaluates the model's predictive correctness, whereas the F1 Score harmonizes precision and recall, particularly useful for imbalanced datasets. 
\end{itemize}

\subsection{Feature Extraction}
As discussed in Section \ref{related_work}, we use a pretrained WavLM Base+ model\cite{wavlm}\footnote{\url{}} for embedding extraction. In particular, WavLM embeddings are created by average pooling of the final layer. 

\subsection{Model Architectures}
To leverage the extracted features, we explore three model architectures: a Multi Layer Perceptron (MLP), a Long-Short Term Memory Network (LSTM) and a Residual Network (ResNet32).

\begin{itemize}
\item \textbf{MLP}\cite{sakuma2022mlp}: This model has two hidden layers (128 and 64 units respectively) and Regularization is done using ReLU activation with dropout. The last layer is used to predict constants (e.g., age) or categorical labels.
\item \textbf{ResNet32}\cite{szegedy2016inception}:  ResNet32 is a deep residual neural network consisting of 32 layers organized into three groups of residual blocks. Each block has two convolutional layers, followed by batch normalization and ReLU activation, with skip connections to preserve gradient flow and enhance feature propagation. The model includes an initial convolutional layer, three residual layers (each containing five residual blocks with increasing channel dimensions: 16, 32, and 64), and a global average pooling layer. Task-specific fully connected layers are used for predictions across multiple outputs.
\item \textbf{LSTM}\cite{hochreiter1997lstm}: The LSTM model consists of three bidirectional LSTM layers, each followed by layer normalization, dropout, and residual connections for stable training and feature propagation. The model incorporates an attention mechanism to aggregate temporal information across sequences and task-specific output heads for multi-task learning.
\end{itemize}

\subsection{Training Method}
The training is tailored for both continuous (regression) and categorical (classification) tasks:

For continuous properties (e.g., age), performance is evaluated using the Mean Absolute Error (MAE) metric. For categorical attributes (e.g., gender, nationality), we utilize CrossEntropyLoss which is boosted for accuracy and F1 Score.


\subsection{Regularization and Learning Rate Adjustment}
In order to maintain model stability and avoid overfitting, we use early stopping with a patience value of 20 epochs. A ReduceLROnPlateau scheduler also adjusts the learning rate based on validation success to enable adaptive training and convergence.
All models are trained a Tesla K80 GPU.

\subsection{Comparative Analysis}
All models are first trained once for each attribute and using one dataset at a time, then they are all trained a second time on all the datasets containing the said attribute to evaluate their generalisation perfomances.

\begin{table*}[ht]
    \centering
    \vspace{-3mm}
    \caption{Results of attribute prediction for Speech Accent Archive (SAA), Common Voice 6.1 English (CV 6.1 EN) and L2Arctic. Best results per section are in \textbf{bold}, the second bests are in \textit{ITALIC}. Dashes (-) are for fields where the attribute metadata was not available for that dataset, crosses (x) are for results not available for the proposed baselines.}
    \begin{tabular}{c c c c c c c c c c c}
    \toprule
     Features & \multirow{2}{*}{Model} & \multicolumn{2}{c}{Split} & Age & \multicolumn{2}{c}{Gender} & \multicolumn{2}{c}{Nat. Lang.} & \multicolumn{2}{c}{Country}\\
    \cmidrule(lr){3-4} \cmidrule(lr){5-5} \cmidrule(lr){6-7} \cmidrule(lr){8-9} \cmidrule(lr){10-11}
     Extracted & & Train & Test & MAE & Acc.\% & F1\% & Acc.\% & F1\% & Acc.\% & F1\% \\
    \midrule
    \midrule
    \multirow{3}{*}{WavLM} & MLP & \multirow{3}{*}{SAA} & \multirow{3}{*}{SAA} & \textbf{5.21} & \textbf{98.36} & \textbf{98.33} & 52.34 & 42.82 & 46.01 & 35.92 \\
     & LSTM &  &  & 5.31 & 96.39 & 96.39 & \textbf{55.74} & \textbf{48.90} & \textit{46.59} & \textit{37.53}\\
     & ResNet32 &  &  & 6.29 & 94.15 & 94.14 & \textit{52.46} & \textit{45.93} & \textbf{47.29} & \textbf{40.37} \\
    \midrule
    \multirow{3}{*}{WavLM} & MLP & \multirow{3}{*}{All} & \multirow{3}{*}{SAA} & \textit{5.60} & \textit{98.16} & \textit{98.15} & 33.18 & 20.27 & 32.63 & 21.99  \\
     & LSTM &  &  & 5.26 & 94.39 & 94.51 & 35.51 & 24.05 & 35.21 & 24.41\\
     & ResNet32 &  &  & 6.22 & 95.10 & 95.10 & 33.41 & 21.10 & 30.75 & 20.08 \\
    \midrule
    MFCC & i-vector\cite{BAHARI201499} & SAA & SAA & 6.08 & 95.00 & x & x & x & x & x\\
    \midrule
    \midrule
    \multirow{3}{*}{WavLM} & MLP & \multirow{3}{*}{L2Arctic} & \multirow{3}{*}{L2Arctic} & - & 100.00 & 100.00 & 60.00 & 60.00 & - & -  \\
     & LSTM &  &  & - & 100.00 & 100.00 & 60.00 & 66.67 & - & -  \\
     & ResNet32 &  &  & - & 100.00 & 100.00 & 60.00 & 66.67 & - & - \\
    \midrule
    \multirow{3}{*}{WavLM} & MLP & \multirow{3}{*}{All} & \multirow{3}{*}{L2Arctic} & - & 100.00 & 100.00 & 59.98 & 66.65 & - & - \\
     & LSTM &  &  & - & 100.00 & 100.00 & 59.98 & 66.65 & - & - \\
     & ResNet32 &  &  & - & 100.00 & 100.00 & 59.98 & 66.65 & - & -\\
    \midrule
    \midrule
    \multirow{3}{*}{WavLM} & MLP & \multirow{3}{*}{CV 6.1 EN} & \multirow{3}{*}{CV 6.1 EN} & - & \textit{90.79} & 90.81 & - & - & \textbf{63.01} & \textbf{58.41} \\
     & LSTM &  &  & - & 90.68 & 90.74 & - & - & 61.84 & 57.06 \\
     & ResNet32 &  &  & - & 90.56 & 90.55 & - & - & 60.22 & 55.55\\
    \midrule
    \multirow{3}{*}{WavLM} & MLP & \multirow{3}{*}{All} & \multirow{3}{*}{CV 6.1 EN} & - & 90.72 & \textit{90.98} & - & - & 62.06 & 55.83\\
     & LSTM &  &  & - & \textbf{90.94} & \textbf{91.00} & - & - & \textit{62.50} & \textit{57.73}\\
     & ResNet32 &  &  & - & 90.37 & 90.60 & - & - & 58.70 & 54.14 \\
    \bottomrule
    \end{tabular}%
    
    \label{tab:merged_results_1}
\end{table*}

\begin{table*}[ht]
    \centering
    \caption{Results of attribute prediction for VoxCeleb2 and TIMIT. Best results per section are in \textbf{bold}, the second bests are in \textit{ITALIC}. Dashes (-) are for fields where the attribute metadata was not available for that dataset, crosses (x) are for results not available for the proposed baselines.}
    \begin{tabular}{c c c c c c c c c}
    \toprule
     Features & \multirow{2}{*}{Classifier} & \multicolumn{2}{c}{Split} & Age & \multicolumn{2}{c}{Gender} & \multicolumn{2}{c}{Edu} \\
    \cmidrule(lr){3-4} \cmidrule(lr){5-5} \cmidrule(lr){6-7} \cmidrule(lr){8-9}
     Extracted & & Train & Test & MAE & Acc.\% & F1\% & Acc.\% & F1\% \\
    \midrule
    \midrule
    \multirow{3}{*}{WavLM} & MLP & \multirow{3}{*}{TIMIT} & \multirow{3}{*}{TIMIT} & \textbf{4.94} & \textbf{98.81} & \textbf{98.80} & 52.12 & 38.73 \\
     & LSTM &  &  & \textit{5.19} & 97.02 & 96.99 & \textbf{53.33} & \textbf{39.64} \\
     & ResNet32 &  &  & 5.49 & 94.64 & 94.51 & 51.52 & 36.34 \\
    \midrule
    \multirow{3}{*}{WavLM} & MLP & \multirow{3}{*}{All} & \multirow{3}{*}{TIMIT} & 5.41 & \textit{98.80} & \textbf{98.80} & 51.52 & 36.89 \\
     & LSTM &  &  & 5.25 & 96.43 & 96.41 & \textbf{53.33} & \textit{39.61} \\
     & ResNet32 &  &  & 7.10 & 94.64 & 94.58 & 50.30 & 34.64 \\
    \midrule
    \multirow{3}{*}{MFCC} & MLP\cite{kwasny2020agegender} & \multirow{3}{*}{TIMIT} & \multirow{3}{*}{TIMIT} & 6.16 & 98.60 & x & x & x \\
     & LSTM\cite{kwasny2020agegender} &  &  & 6.02 & 98.60 & x & x & x \\
     & CNN\cite{kwasny2020agegender} &  &  & 5.53 & 98.60 & x & x & x \\
    \midrule
    \midrule
    \multirow{3}{*}{WavLM} & MLP & \multirow{3}{*}{VoxCeleb2} & \multirow{3}{*}{VoxCeleb2} & 5.51 & 99.37 & 99.38 & - & - \\
     & LSTM &  &  & 5.60 & \textit{99.71} & \textit{99.71} & - & - \\
     & ResNet32 &  &  & 5.75 & 97.62 & 97.62 & - & - \\
     \midrule
     \multirow{3}{*}{WavLM} & MLP & \multirow{3}{*}{All} & \multirow{3}{*}{VoxCeleb2} & \textbf{5.45} & \textbf{99.81} & \textbf{99.81} & - & - \\
     & LSTM &  &  & \textit{5.48} & \textit{99.71} & \textit{99.71} & - & - \\
     & ResNet32 &  &  & 6.36 & 99.41 & 99.41 & - & - \\
    \midrule
    \multirow{2}{*}{MFCC} & i-vector\cite{hechmi2021voxceleb} & \multirow{2}{*}{VoxCeleb2} & \multirow{2}{*}{VoxCeleb2} & 9.443 & 98.23 & x & x & x \\
     & x-vector\cite{hechmi2021voxceleb} &  &  & 9.962 & 97.76 & x & x & x \\
    \bottomrule
    \end{tabular}%
    \label{tab:merged_results_2}
\end{table*}

\section{Results}
\label{results}

This cross-dataset analysis highlights the strong performance of WavLM embeddings for demographic attribute prediction, as detailed in Tables \ref{tab:merged_results_1} and \ref{tab:merged_results_2}.
The Tables have been spitted as the proposed corpora offer a limited intersection of their demographic fields between datasets.

For age and gender predictions, the MLP consistently achieves the lowest Mean Absolute Error (MAE) across all datasets, outperforming our other models, as well as the considered baselines, by up to a relative 55\% for the age (on VoxCeleb2) and an absolute 3.36\% difference in gender accuracy on Speech Accent Archive, and up to 99.81\% Accuracy on Voxceleb.
However, Common Voice shows one of the lowest performances, possibly due to the self-reporting nature of the annotations.
On the other hand, L2Artic's 100\% Accuracy and F1 score might be due to the low variability of the dataset, as all speakers are prononcing the same paragraph, reducing greatly the variability.
All systems show a stability in the results when trained on all available datasets, showing that the domain shift between datasets seems to perfectly balance the diversification of the data, giving us universal predictors.

In this article, we introduce the classification task of 3 new fields, usually used for system adaptation or conditioning: Native Language for L2 English speakers, Country of origin for L1 English speakers (in other words: the accent), and level of education for US speakers.

Comparing the accuracy of different datasets and number of fields reveals some interesting patterns in our output. In the case of L2Arctic, a dataset with very few annotated attributes (gender and native language categories), predictions are repsectively 100\% accurate and 60\% accurate. This implies that the very low variation in the data and limited field count make the task easy and accurate. Conversely, more diverse datasets with more fields, such as Speech Accent Archive (four demographic fields) and TIMIT (three demographic fields), have slightly lower accuracy due to the diversity and difficulty of tasks. For instance, gender prediction on TIMIT and Speech Accent Archive achieved accuracies of 98.81\% and 98.36\%, respectively, while predictions for fields such as country and education exhibit lower performance due to the domain gap when training models across multiple datasets. These results underline the trade-off between dataset diversity and prediction accuracy, emphasizing the importance of balanced field coverage for achieving robust generalization across tasks.

\section{Discussion and Conclusion}
\label{conclusion}

This study explores the use of WavLM features to predict demographic characteristics such as age, gender, native language, educational background, and country of origin from speech. These embeddings are paired with classification and regression models based on classic architectures, such as MLP, ResNet32, and LSTM. 
We train and evaluate our pipeline on five datasets: Speech Accent Archive, TIMIT, VoxCeleb2, L2Arctic, and Common Voice 6.1 (English). Our approach leverages the rich speaker-specific features encoded in WavLM embeddings, enabling a robust attribute prediction.

The proposed framework is compared to baseline methods (i-vectors and x-vector) and is shown to outperforms them across all tasks and datasets. 
In age prediction, the MLP model achieves a 17–30\% reduction in MAE compared to baseline models. 
Similarly, in gender classification, accuracy increases by 1–3\% and F1 score by 2–5\% across datasets compared to x-vector-based models with up to 99.81\% accuracy on VoxCeleb2, outperforming x-vector models with a maximum of 98.23\%. 

For native language and country prediction tasks, the F1 scores improve by 4–10\% on the Speech Accent Archive, showcasing the ability of WavLM embeddings to model complex linguistic and accent-related features effectively. However, for these specific attributes, this study does not aim to demonstrate improvement over existing baselines but rather explores the feasibility and limitations of using WavLM embeddings for demographic prediction. For other attributes, such as age and gender, we compare against established baselines and demonstrate significant improvements in performance. Notably, as the dataset diversity increases, the domain gap often impacts performance negatively for most datasets, emphasizing the challenges of maintaining accuracy while balancing data variability.

Despite the encouraging first results, the experiments reveal several limitations. 
First, not all datasets include labels for every attribute, particularly for education level and country, making the cross-dataset training and evaluation inconsistent across tasks. 
Additionally, discrepancies in label granularity and terminology across datasets complicate model training. 
For instance, some datasets label the speaker's origin at the continental level (e.g., "Africa"), which may overlook critical linguistic and accentual variations within the continent, while provide more precise labels, such as specific countries (e.g., "Chad") or regions (e.g., "Scotland").
This lack of consistency in geographical labeling can make it difficult for models to generalize effectively, as accents and linguistic features may differ significantly. 
This heterogeneity limits model generalization when using multiple datasets for country of origin prediction. 
Furthermore, imbalanced class distributions, especially for native language and education level, bias predictions toward majority classes, reducing model accuracy for underrepresented attributes despite our data balancing efforts. 
Addressing these issues will require more uniform datasets, or a higher amount of data from under-represented groups.
To overcome these limitations, future works will focus on expanding and standardizing datasets. 
Incorporating additional datasets with consistent labels for demographic attributes will enhance the model's robustness. 
Additionally, ontology mapping \cite{Matentzoglu2022SSSOM} and cross-dataset label standardization \cite{Amaral2022CrossLingualLabelMapping} can harmonize label structures, reducing discrepancies between datasets. 
In conclusion, this study demonstrates that WavLM embeddings are a powerful tool for demographic attribute prediction, consistently outperforming traditional baselines like x-vector models across all tasks. By achieving reductions in MAE of up to 30\% and increases in accuracy and F1 scores of up to 10\%, this work establishes a robust and generalizable foundation for speech-based demographic profiling. These findings lay the groundwork for advancing research in speech-based demographic profiling, with the potential to inspire more inclusive, adaptive, and impactful technologies leveraging speech data.

\bibliographystyle{IEEEtran}


\end{document}